\documentclass[journal]{IEEEtran}

\usepackage{graphicx}
\usepackage{subcaption}
\usepackage{amsmath,amssymb,amsfonts}
\usepackage{amssymb,amsmath,cite,graphicx,graphics,enumerate}
\usepackage{algorithmic}
\usepackage{multicol}
\usepackage{multirow}
\usepackage{color,soul}

\ifCLASSINFOpdf
\else
\fi

\begin{document}

\title{Deep Reinforcement Learning based Automatic Exploration for Navigation in Unknown Environment}
%
%

\author{Haoran~Li,
        Qichao~Zhang,
        Dongbin~Zhao \emph{Senior Member, IEEE}  
\thanks{
This work is supported by the Beijing Science and Technology Plan under Grants Z181100004618003, and the National Natural Science Foundation of China (NSFC) under Grants No. 61573353, 61803371, 61533017 and 61603268.
This work also supported by Noah's Ark Lab, Huawei Technologies.

Haoran Li, Qichao Zhang, and Dongbin Zhao are with the state Key Laboratory of Management and Control for Complex Systems, Institute of Automation, Chinese Academy of Sciences, Beijing, 100190, China, and also with the University of Chinese Academy of
Sciences, Beijing 100049, China (email : lihaoran2015@ia.ac.cn, zhangqichao2014@ia.ac.cn, dongbin.zhao@ia.ac.cn).}
}

\maketitle

\begin{abstract}
This paper investigates the automatic exploration problem under the unknown environment,  which is the key point of applying the robotic system to some social tasks. The solution to this problem via stacking decision rules is impossible to cover various environments and sensor properties. Learning based control methods are adaptive for these scenarios. However, these methods are damaged by low learning efficiency and awkward transferability from simulation to reality. In this paper, we construct a general exploration framework via decomposing the exploration process into the decision, planning, and mapping modules, which increases the modularity of the robotic
system. Based on this framework, we propose a deep reinforcement learning based decision algorithm which uses a deep neural network to learning exploration strategy from the partial map. The results show that this proposed algorithm has better learning efficiency and adaptability for unknown environments. In addition, we conduct the experiments on the physical robot, and the results suggest that the learned policy can be well transfered from simulation to the real robot.
\end{abstract}


%
\IEEEpeerreviewmaketitle

\section{Introduction}
%
%
%
%
\IEEEPARstart{A}{}s an important subproblem of the robot autonomous navigation, automatic exploration means that the robots move ceaselessly to build the entire environmental map in a new environment without any priori knowledge, which is a hot topic in the field of robotics. It has a wide range of scenario applications in practice, such as the search work of rescue robots and the sweeping work of the robot sweepers in the unknown environment. As the robot has to complete the exploration and mapping task in the completely different and unknown environment, the automatic exploration can also reflect the adaptability of the robotic system.

With the spreading attention from various fields, some noticeable works have been yielded such as frontier-based\cite{yamauchi1997frontier} and information-based\cite{bourgault2002information} methods. The frontier-based method was to decide the next move of the robot via searching the frontier points which were between free points and unknown points. And the information-based method applied Shannon entropy to describe the uncertainty of the environmental map and construct optimization problems, in which way the optimal control variable of the robot can be gained during the automatic exploration process. Unfortunately, in most cases, there is no mathematical optimal solution in the optimization problems constructed by this method. Then, some researchers combined these two methods\cite{gonzalez2002navigation}. This kind of approach employed Shannon entropy to evaluate candidate points selected by frontier and then obtained the optimum point in the next move. Furthermore, to make the exploration more efficient and the map more precise, an extra target item was added into the objective function of the optimization problems\cite{julia2012comparison}. However, the computational burden to solve these multi-objective optimization problems will increase rapidly with the increase of the exploration area. In general, the key to the traditional methods mentioned above is to find the optimum next point according to the current point and the explored map. However, the adaptability is decreased due to the key exploration strategy which is relied heavily on the expert feature of maps. And it is also difficult to design a general terminal mechanism to balance the exploration efficiency and computational burden.

Gradually, learning based exploration methods are considered to tackle the problems above. Krizhevsky \emph{et al.}\cite{krizhevsky2012imagenet} adopted deep convolutional neural networks(CNN) and  designed an efficient classification network for large-scale data, which triggered a wave of research on neural networks. CNN has spurred a lot of big breakthroughs in the image recognition area\cite{lecun2015deep}. Meanwhile, reinforcement learning(RL) has also achieved certain development\cite{mnih2015human}. Since Mnih \emph{et al.}\cite{mnih2015human} applied a deep reinforcement learning(DRL) algorithm and gained better performance than human players in the video game like Atari, the algorithm combining with deep learning(DL) and RL became one of the effective tools for complex decision-making problems. Based on Q-learning\cite{sutton1998reinforcement} and the strong feature presentation ability of CNN, Deep Q-network(DQN) has shown its tremendous potential in the robot control and game decision making. In robot control field, the DRL methods in continuous action spaces can establish the mapping from image inputs to the control policy which are concise\cite{lillicrap2015continuous}\cite{kiumarsi2018optimal}. It has many applications in manipulator controling\cite{zhu2017event}, intelligent driving\cite{zhao2017model} and games\cite{zhu2017iterative}\cite{zhao2016experience}. In game decision making, the DRL methods in discrete action spaces, such as AlphaGo\cite{silver2016mastering}, show the strong search capability within the high-dimensional decision space. Considering these merits of DRL, some researchers attempt to apply it to the exploration and navigation of intelligent agents in unknown environment.

Currently, DL or DRL-based methods are mostly adapted in the robotics visual control or navigation tasks, which build the mapping relationship between raw sensor data and control policy. In this way, the intelligent agents can move autonomously relying on the sensor data. However, there are two disadvantages for this end-to-end approach. On the one hand, the information provided by mature navigation methods in robotics is ignored, which means the intelligent agents need to learn how to move and explore effectively from raw data. It definitely increases the training difficulty of intelligent agents. On the other hand, the reality gap between the synthetic and real sensory data imposes the major challenge from simulation to reality. To the best of our knowledge, the reality gap for automatic exploration based on DRL in unknown environment are not addressed in the existing work.

Motivated by these, we construct an automatic exploration framework based on map building, decision making and planning modules. This framework makes it easy to combine the fairly complete map building and navigation methods in robotics. Since the visual feature of the built grid map could guide the robot to explore unknown environment efficiently, we propose a DRL-based decision method to select next target location with the grid map of the partial environment as input. Compared with some existing literature, the main contributions emphasize in two parts.
\begin{enumerate}
  \item  In contrast to the end-to-end approach with the raw sensor data as input and control policy as output, our exploration framework combing with traditional navigation approaches can bridge the reality gap from the simulation to physical robots. Meanwhile, the training difficulty is reduced with the proposed framework.

  \item  Compared with traditional methods\cite{gonzalez2002navigation}\cite{julia2012comparison}, we design a value network of deep reinforcement learning and combine the DRL-based decision algorithm with classical robotic methods, which can improve the exploration efficiency and adaptability in unknown environment.

 \end{enumerate}

This paper is organized as follow. Section II describes related works in robot automatic exploration and the intelligent agent navigation with DRL. Section III presents our definition and analysis about the robot automatic exploration. Then, we give the details of our proposed algorithm and network architecture in Section IV. Furthermore, the simulation experiment and the practicality experiment are displayed and the comparison among different methods is given. Finally, in section V, we summarize our work and discuss future work.

\section{Related Work}
In general, the automatic exploration scheme can be divided into two categories: traditional methods and intelligent methods. The former, which have been attracting considerable interest since the 1990s, employs expert features of maps for selecting the next goal point. The latter learns the control or strategy directly from the observation based on the learning approaches.

The widely-used traditional method for selecting exploration points was suggested by Yamauchi\cite{yamauchi1997frontier}. This method detected the edges between free space and unexplored space based on the occupancy grid maps and subsequently calculated the central point of each edge as the next candidate point. Then, the robot moved to the nearest point by performing the depth-first-search algorithm. However, the safety of the next point is uncertain, since the distance between the point and obstacle may be too small. Gonz¨¢lez-Banos \emph{et al.}\cite{gonzalez2002navigation} defined the safety regions, and built a Next-Best-View algorithm to decide which region the robot should move to. The simultaneous localization and mapping(SLAM) algorithm and path planning were embedded in this algorithm, where the former was applied to reconstruct the environment and the latter was utilized for generating the actions of the robot.

Another typical traditional methods were connected with information theory. Bourgault \emph{et al.}\cite{bourgault2002information} took advantage of Shannon entropy to evaluate the uncertainty of the robotic positions and the occupancy grid maps, and exploited this information metric as the utility of the robot control. The control was derived from a multi-objective problem based on the utility. The utility function of localization was stemmed from the filter-based SLAM algorithms\cite{stentz2003fastslam}. There is a large volume of existing studies describing the role of the graph-based SLAM algorithm\cite{grisetti2010tutorial}. Different from filter-based algorithms, the graph-based methods proposed pose graph which applied poses of robots and landmarks as nodes and exploited the control and measurements as edges. Note that new pose caused by robot movement influenced the covariance matrix of the pose graph. To evaluate the quality of effect from the new pose, Viorela \emph{et al.}\cite{ila2010information} defined a mutual information gain for a new edge in the pose graph. With the rapidly-exploring random tree, they searched the paths to explore the unknown environment and used the information gain to evaluate the path entropy, afterward they choose the one which minimized the map entropy.

Due to the significant recent development in DRL, a number of researchers in intelligent control area attempted to regard the robot exploration as an optimal control problem constructed by information-based methods and employ reinforcement learning to settle the sequence decision problem.
Martinez \emph{et al.}\cite{julia2012comparison} modeled the exploration as the partial observation Markov decision process(POMDP)\cite{monahan1982state} and employed the direct policy search to solve the POMDP. For the great perception ability of DL, it has been widely employed to build powerful methods for learning the mapping from sensor data to the robot control. In 2016, Tai \emph{et al.}\cite{tai2016mobile} proposed a perception network with RGB-D image inputs and a control network trained by the DQN. Note that this algorithm only ensures the robot wandering without collision. Bai \emph{et al.}\cite{bai2017toward} constructed a supervised learning problem to reduce Shannon entropy of the map. They fed a local pitch of the map into neural networks and predicted which direction the robot should move. This algorithm often fell into a corner since it only perceived the local environment which was completely explored.

Much of the current literature on intelligent methods pays a particular attention to game agents navigation and exploration in a virtual environment.  Mirowski\cite{mirowski2016learning} trained an Asynchronous Advantage Actor-Critic(A3C) agent in a 3D maze. They embedded long short-term memory(LSTM) into the agent to give it the memory. To improve learning efficiency, they preceded a depth prediction task and a loop closure detection task. This attempt presented a remarkable performance in this virtual environment. Compared to adding external tasks, Pathak proposed an intrinsic curiosity module(ICM)\cite{pathak2017curiosity}. The deep neural network was utilized to encode state features and constructed a world model applied to predict the next step observation in feature space. This model was used to predict the next step observation. The differences between the next step encoder features and the prediction of next step observation features by the world model were employed to drive the agent to move around. Then Oleksii \emph{et al.}\cite{zhelo2018curiosity} applied this approach in robot navigation. They combined policy entropy and A3C to calculated robot control.

\section{Problem Statement}
In order to achieve the autonomous navigation, the robot receives sensor data and simultaneously builds surroundings map during moving in the unknown environment. To create a more similar map to the actual situation, the robot exploration algorithm is essential. Under traditional navigation frameworks, the robot has to generate the environment map and locate itself. With the development of DL and DRL, there are several end-to-end control methods by DRL over the years. Those methods feed the sensor data and surroundings as the input of deep neural networks and output the control for the robot.

A momentous advantage of end-to-end methods is concise. However, there is a wide gap to apply the technique to a real robot. The end-to-end learning methods have to conduct massive trial-and-error experiences, so most of the researchers put this learning process in the simulation environment. Consequently, this behavior raises three problems: 1) The algorithm needs a long time to converge since the robot starts from learning movement to explore the unknown environment. 2) Since the real environment is not the exact same with the simulation environment, there is no guarantee that the mapping represented by deep neural network learned under simulation environment has a superior generalization under the real environment. 3) The mechanism and sensor errors of the physical robot may cause that the intelligent control law is not applicable to the physical robot.

\begin{figure}
  \centering
  \includegraphics[width=0.48\textwidth]{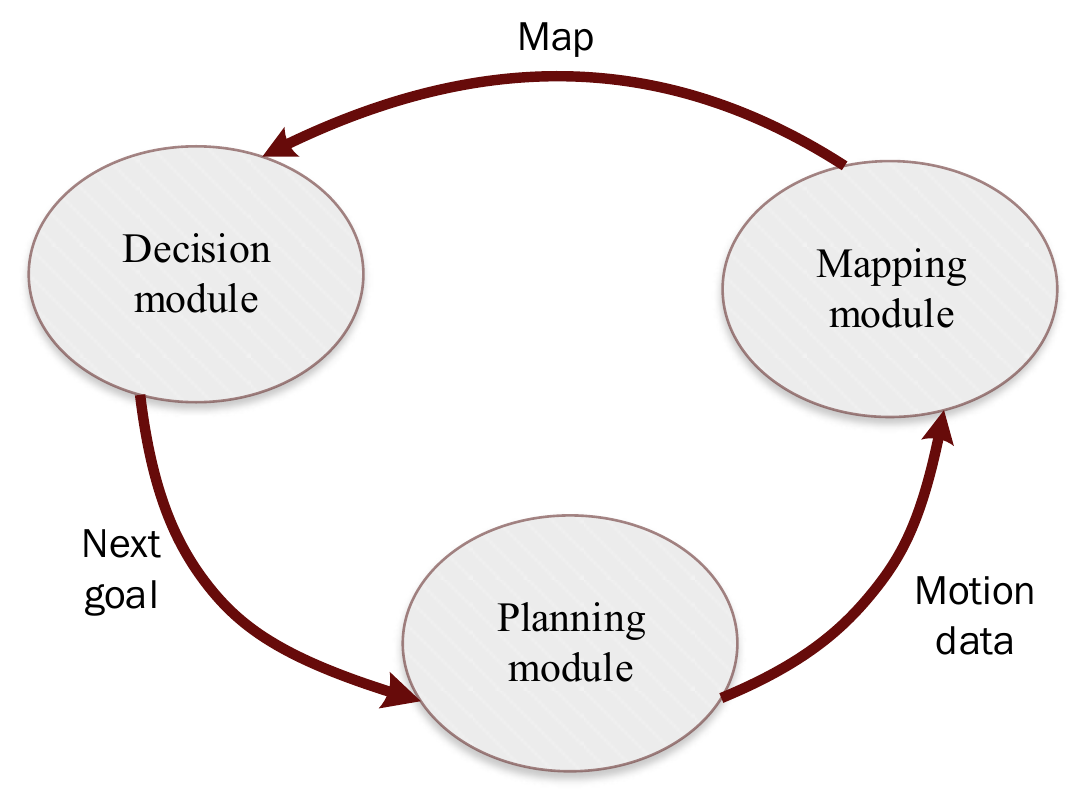}
  \caption{The relationship between the modules of the exploration framework. Decision module receives data from the mapping module and sends the next goal point to planning module. Planning module calculates the potential trajectories from the current position to the goal position and sends the control data to the mapping module.}
  \label{Fig_system}
\end{figure}

Based on the above issues, we separate the robot exploration into three modules: decision module, mapping module, and planning module. Therefore, We can design the planning algorithms for different robots with the same decision module since each module is independent of each other. The planning algorithms guarantee the robots can move safely without learning in a known environment. With the help of the traditional navigation algorithms, the decision module focuses on the efficient exploration strategy. Figure \ref{Fig_system} describes the relationship between the three modules. In the following, we form each module more general for extensions and unfold each module to explain its function.
\subsection{Decision module}
According to the historical observation, this module chooses the goal point where the robot should move to. In this paper, the module input is the map of surroundings built by the mapping module. The decision module is modeled as
\begin{equation}
g_t = f_{decision} (l_{0:t}, m_t),
\end{equation}
where $g_t$ is the goal point, $l_{0:t}$ are the robot positions from time 0 to $t$, and $m_t$ is the built surroundings map at time $t$. There are several efforts focused on figuring out the function $f_{decision}$, most methods are based on the frontier methods and with the heuristic search. Those methods are required to establish massive rules for complex environments, which results in heavy dependencies on expertise.

In this paper, we will mainly focus on finding out the more concise and the general function $f_{decision}$. We propose a DRL-based decision method in the next section. Compared to the frontier-based method, our method learns the visual feature for exploration automatically. Instead of stacking the rules for exploration, our method learns the feasible strategy from trial and error and cover more complex scenes.

\subsection{Planning module}
This module aims to ascertain the feasible trajectory from the current position to the goal point. The planning module is modeled as
\begin{equation}
\tau_{t:t+T} = f_{planning}(l_{0:t}, g, m_t),
\end{equation}
where $\tau_{t:t+T}$ is the planning trajectory for robot control. The widely-investigated framework for many planning approaches which implement the planning function includes the global planning method and the local planning method. The aim of the former is to determine the shortest path from the current position to the goal position on the map, and the latter attempts to convert the path to a trajectory and adjust the trajectory based on the real-time sensor data. It should be mentioned that recent studies related to the planning module have shown different frameworks.  Those particular methods modeled as deep neural networks take the sensor data and goal point as the input and build the mapping from the input to the control.

Considering that $A^*$\cite{yao2010path} is a simple and efficient search method, we apply $A^*$ as the global planner. The results of $A^*$ search is a path composed of the discrete points. And the path cannot control the robot directly. We employ timed-elastic-band(TEB) method\cite{rosmann2017kinodynamic} as the local planner which convert the path to the robot velocity. Compared to the pure tracking, this method can avoid obstacles which are not in the built map(such as moving objects) according to the real-time point cloud from light detection and ranging(LiDAR).

\subsection{Mapping module}
Mapping module processes the sensor data in sequence and produces the robot pose and surroundings map at each timestamp, which is also known as SLAM. The function of the SLAM can be written as
\begin{equation}
m_t, l_t = f_{mapping}(l_{t-1}, m_{t-1}, u_t, z_t).
\end{equation}
Here $u_t$ and $z_t$ are the control and observation at time $t$, respectively. The major SLAM methods can be divided into two categories: filter-based methods and graph-based methods. Filter-based methods make use of Kalman filter or expand Kalman filter to update the estimation of the robot and landmark(map) pose. Graph-based methods collect the control and the observation to construct a pose graph and take advantage of nonlinear optimization to estimate the robot pose. Moreover, Parisotto\cite{parisotto2017neural} proposed a neural SLAM method which exploited deep neural network fitting the function $f_{mapping}$.

We utilize a graph-based method named Karto SLAM. It is simple to implement and works well in most scenes. Compared to the filter-based method, the graph-based method can diminish the error and obtain the extract solution consistently. Therefore, more and more papers focus on graph-based methods recently\cite{grisetti2010tutorial}.

In this paper, we place great emphasis on constructing the DRL algorithm for the decision module to implement efficient exploration. Compared with traditional decision methods, the DRL-based decision algorithm can eliminate trivial details of complicated rules. Different from end-to-end methods, the approaches based on this framework have a higher degree of modularity, and it is more flexible and interpretable.

\section{DRL-Based Decision Method}

\subsection{Objective function}
The objective of the robot exploration is to build an environment map which is most similar to the real environment. Considering the battery capacity, the robot is designed to  select a shorter path to explore the environment. Then, we give the following objective function
\begin{equation}
c (\hat{m}, x_{t=0:T}) = \min_{u_{t=0:T}} \| m - \hat{m} \|^2 + L(x_{t=0:T}).
\end{equation}
Here $\hat{m}$ is the estimation map, $m$ is the real map, and $L( 	\cdot )$ is the path length during exploration from $t = 0$ to $t = T$.

However, since the actual map is non-available due to the unknown environment, it is hard to handle the objective function and find the minimum solution. Inspired by the Shannon entropy for evaluation of the exploration\cite{bourgault2002information}, we exploit the occupancy grid map to represent the environment. Note that $p(m_{i, j})$ is the occupied probability of the cross grid of $i$-th column and $j$-th row. Each grid has three possible states: unknown, free and occupied. And the entropy of the map is
\begin{equation}
H(m) = - \sum_i \sum_j p(m_{i,j})\text{log} p(m_{i, j}).
\end{equation}
This objective function  evaluates the uncertainty of the built map.

\subsection{Deep Reinforcement Learning}
We take exploration as a sequence decision-making task. A Markov decision process defined as a tuple  $< S, A, T, R, \gamma >$ is a framework for decision making. $S$ denotes a finite set of states. $A$ is the set of the actions that the agent can take. $T(s'|s,a)$ is the transform distribution over the next state $s'$ given the agent took the action $a$ with the state $s$. $R(s, a)$ is the reward the agent receives after taking action $a$ in state $s$. $\gamma$ is the discount factor. The goal of the agent is to maximize the expectation of long-term cumulative reward $V(s)$

\begin{equation}
V(s) = \max \sum_{t=1}^T E_{(s_t, a_t) \sim p_{\theta}(s_t, a_t)} [r(s_t, a_t)].
\end{equation}
RL is an intelligent method for robot control\cite{zhang2018event}. Q-learning is a popular method to learn an optimal policy from samples generated from interactions with the environment. The key to Q-learning is calculating the Q-value function which evaluates the action took in the state. And the optimal Q-value obeys the Bellman optimal equation as follows
\begin{equation}
Q^*(s, a) = E_{s'}[r + \gamma \max_{a'} Q^* (s', a') | s, a].
\end{equation}
Since this optimization problem is nonlinear, it is difficult to obtain the analytic solution. The common iterative method for this problem is the value iteration as follows:
\begin{align}
Q(S_t, A_t) \leftarrow &Q(S_t, A_t) + \alpha [R_{t+1} + \gamma \max_{a' \in A} Q(S_{t+1}, a')\nonumber \\ &-Q(S_t, A_t)].
\end{align}
Note that recent advances in DL have facilitated investigation of feature extraction. Mnih \emph{et al.}\cite{mnih2015human} took a deep neural network as a function approximation for Q-value function and trained the networks by minimizing the mean squared error(MSE)
\begin{equation}
L^{DQN} = E [(R_{t+1} + \max_{a' \in A} Q(S_{t+1}, a'; \theta^-) - Q(S_t, A_t;\theta))^2],
\end{equation}
where $(S_t, A_t, R_{t+1}, S_{t+1})$ are sampled from the experience replay buffer collected by the agent. This DQN algorithm achieved a huge success in many games such as Go, Atari, etc.

Recently, many researchers introduce DQN algorithm for robot navigation. Most of them apply the original sensor data as the algorithm inputs, such as point clouds of LiDAR\cite{chen2017decentralized}, images\cite{li2017torcs}\cite{chen2015deepdriving}\cite{zhao2017deep}\cite{chen2018multi}. With the powerful representation of deep neural networks, these methods can extract key features without expert experiences. Nevertheless, compared to the empirical features, the original sensor data has a higher dimension which increases state space sharply. From this perspective, the agent has to take massive trial and error to cover all the potential state-action combination, which takes a long time. In view of the safety and lifetime of the real robots, it is impossible to take the extensive trials in the real environment. Hence, most researches trained the agent in a simulated environment. However, there are two common problems in transferring the virtual agent into a real agent. First, since the error functions of the sensors are different between virtual and real environment, It is hard to guarantee the good generalization of the DRL algorithm from virtual to real. Second, due to the mechanical error, it is uncertain that the real robots will have an excellent performance with the control law learned in a virtual environment.

As discussed above, we introduce the hierarchy decision framework in the last section. We separate the decision module from the planning module. Decision module provides the next goal point. And planning module plans the path from the current position to the goal point. In this way, the control policy derived from the traditional optimal control has a good match with the real robot.

In this section, we aim to design a novel DRL algorithm for decision module. The input of the decision module is a map, current and historical robot positions obtained from the mapping module. We obtain the optimal decision sequence
\begin{align}
x^*_{t=0:T} &= arg\min H(m_T) + L(x_{t=0:T})\nonumber \\
           &= arg\min H(m_T)-H(m_0) + L(x_{t=0:T}\nonumber) \\
           &= arg\min \sum_{t=1}^T [H(m_t)-H(m_{t-1})] + \sum_{t=1}^T L(x_{t-1}, x_t)\nonumber \\
           &= arg\max \sum_{t=1}^T [H(m_{t-1})-H(m_{t}) - L(x_{t-1}, x_t)].
\end{align}
Therefore, we can define the reward function for exploration as follows
\begin{equation}
r_t = \alpha (H(m_{t-1})-H(m_{t}) - L(x_{t-1}, x_t)),
\end{equation}
where $\alpha$ is coefficient to make the training process more stable.

In view of the safety, we attach the heuristic reward function such as
\[ r_t =
  \begin{cases}
    -1       & \quad \text{if the next point is in unknown space,}\\
    -1       & \quad \text{if the next point is too close to the obstacle.}
  \end{cases}
\]

To stop the exploration in time, we introduce the terminal action and define the reward for this action
\[ r_t =
 \begin{cases}
 1          & \text{if the ratio of explored region in map $\rho$} > 0.85, \\
 -1         & \text{otherwise.}
 \end{cases}
\]
\textbf{{Remark 1:}} The target of automatic exploration is to build the map of the unknown environment based on the navigation and mapping technique. Note that the logic of reward signal is designed based on the safety and efficiency. As the robot cannot confirm the dangerous status of the unknown space which may be the obstacle, wall and so on. Therefore, a penalty signal is given to encourage the robot to choose next goal point in the free space. Besides, we hope the distance from the goal point to the obstacles will be larger than the robot geometrical radius in consideration of collision avoidance. Combined with the designed efficient exploration logic, the robot can find out the safety and efficient policy in unknown environment. It should be mentioned that the ratio $\rho$ is a hyperparameter to balance the explored region rate and explored efficiency. Here we choose it as 0.85 based on  experiments

\begin{figure}
  \centering
  \includegraphics[width=0.48\textwidth]{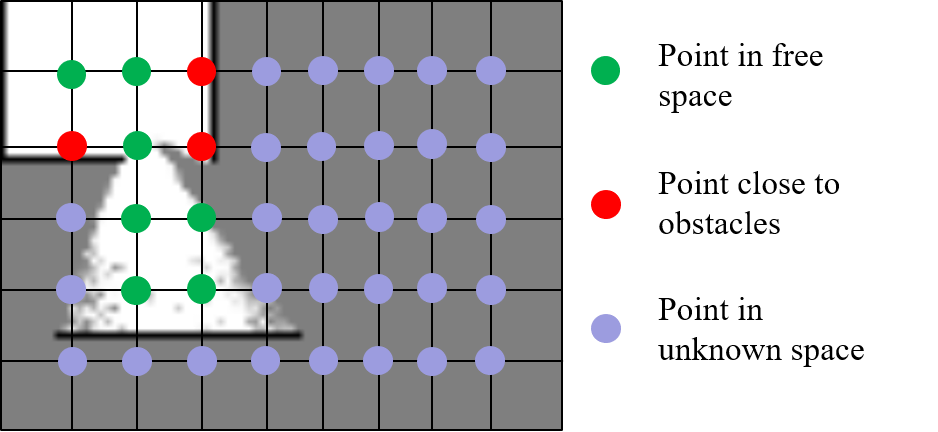}
  \caption{The action space based on the occupancy grid map. In the grid map, the white and gray area is the free and unknown space, respectively. The black edges are the occupied grids which are the contours of the obstacle. The points are the action positions which are sampled regularly from the grid map. The green points are safe actions, and others are dangerous.}
  \label{Fig_og_sp}
\end{figure}

\begin{figure*}
\centering
  \includegraphics[width=1.0\textwidth]{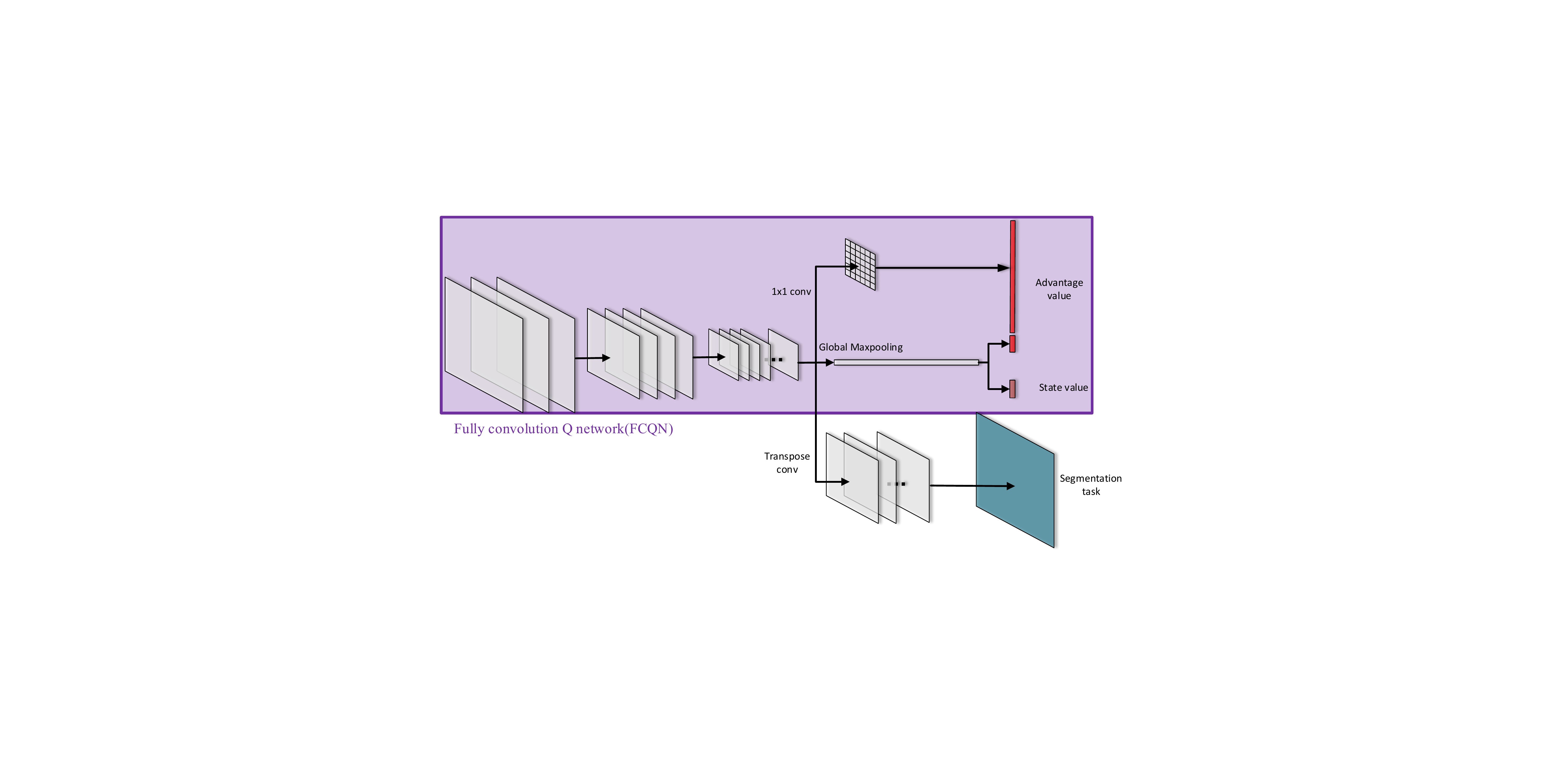}
  \caption{Fully convolutional Q-network with an auxiliary task. The lower layers of this network are convolutional as in the original DQNs.}
  \label{Fig_net}
\end{figure*}

\subsection{Action Space and Network Architecture}
The action of exploration is selecting a point from the built map. Since we utilize the occupancy grid map, the action space is discrete. To simplify the problem and reduce the searching space, we exploit rasterized sampling on the grid map. And the action space is composed of these sampling points. Figure \ref{Fig_og_sp} describes the occupancy grid map and the action space for the proposed algorithm.

Since the points in action space are obtained from regularly sampling, we can design a convolutional neural network to estimate the value of each point in the map. The designed network architecture is shown in Figure \ref{Fig_net}. The lower layers are encoder layers where the output dimension is the same as action space.

To estimate the Q-values, we construct a network named fully convolutional Q-network(FCQN) based on two convolutional branches after encoder layers. The one applies convolution to calculate score map which has the same size with encoder feature maps. Each element of the score map represents the advantage function of the corresponding point in action space. The other exploits max-pooling operation to collect the feature vector of the whole map. Subsequently, the fully connection layer estimates the advantage value of the terminal action and the state value. The Q-value of each point action is derived from \cite{wang2015dueling}
\begin{equation}
Q(s, a; \alpha, \beta) = V(s; \beta) + (A(s, a; \alpha) - \frac{1}{|A|} \sum_{a'} A(s, a'; \beta)).
\end{equation}
The terminal action Q value is
\begin{equation}
Q(s, a; \beta) = V(s; \beta) + (A(s, a; \beta) - \frac{1}{|A|} \sum_{a'} A(s, a'; \beta)).
\end{equation}

Compared with original DQNs architecture, our network makes several significant improvements. One the one hand, the FCQN is more accessible to train and has less overfitting risk due to the fewer parameters. The original DQNs employs fully connected layer after flattening feature maps and introduces massive trainable parameters. By contrast, the FCQN only has several convolutional kernel parameters. The fully connected layer in original DQNs needs the inputs to have the same dimension, which restricts its adaptability for various size environment. On the other hand, since the fully convolutional network accepts the different size inputs, the FCQN has better adaptability for changeable map size. Those properties make our network more convenient to train in a new environment continuously.

\subsection{Auxiliary task}
Compared with the empirical features, putting the map image as the input of DRL increases the search space rapidly. The agent needs more time to analyze every possible state-action combination, which increased difficulties in the training process.

Some researchers proposed auxiliary tasks by enhancing prior knowledge to improve training efficiency. Mirowski \emph{et al.}\cite{mirowski2016learning} implemented depth prediction and loop closure prediction as the auxiliary task to improve the performance and training efficiency of DRL algorithms.  Lei \emph{et al.}\cite{tai2016mobile} trained the deep convolutional network with classification task and then exploited this network to extract the features for the RL. The feature input drastically reduces the state dimension and exploration time instead of the original image input. Unfortunately, these features are usually not the task-oriented, there is no guarantee for the quality of the features.

Considering the crucial roles of the map edge in navigation and exploration, we present the edge segmentation as the auxiliary task for FCQN. Note that the edges composed of two elements. One is the contour of obstacles which is essential for avoiding the collision during navigation.  The other is the boundary between free space and unknown space. This boundary is named frontier employed to produces the candidate points.

The third branch for segmentation appended in the network architecture as shown in Figure \ref{Fig_net}. And we call the whole network with segmentation branch as AFCQN (fully convolutional Q-network with an auxiliary task). We apply  the decoder block which consists of twice deconvolution after lower feature maps. The output of this branch has the same dimension as the input map. The loss function for this branch is
\begin{equation}
L_s = -\sum_p \sum_{c=1}^M y_{o, c} log p_{o, c},
\end{equation}
where $p$ represents the pixel in the map, and $M$ is the set of the possible classes. In this case, $M$ includes 3 types such as the contours of the obstacle, frontiers, and others. $y_{o, c}$ is the binary indicator if the class label $c$ is the correct classification for observation $o$. And $p_{o, c}$ means that the predicted probability  observation $o$ is of the class $c$.

Therefore, the lower convolutional layers update the parameters by two-part back-propagation error.
\begin{equation}
\Delta w = - \epsilon (\frac{\partial L^{DQN}}{\partial w} - \lambda \frac{\partial L_s}{\partial w}),
\end{equation}
where $\epsilon$ and $\lambda$ are the learning rate and the balance parameter for the segmentation task, respectively. $w$ is the parameter of lower convolutional layers. These analytical procedures and the results are described in the next chapter.

\section{Whole System}
\begin{figure}
  \centering
  \includegraphics[width=0.48\textwidth]{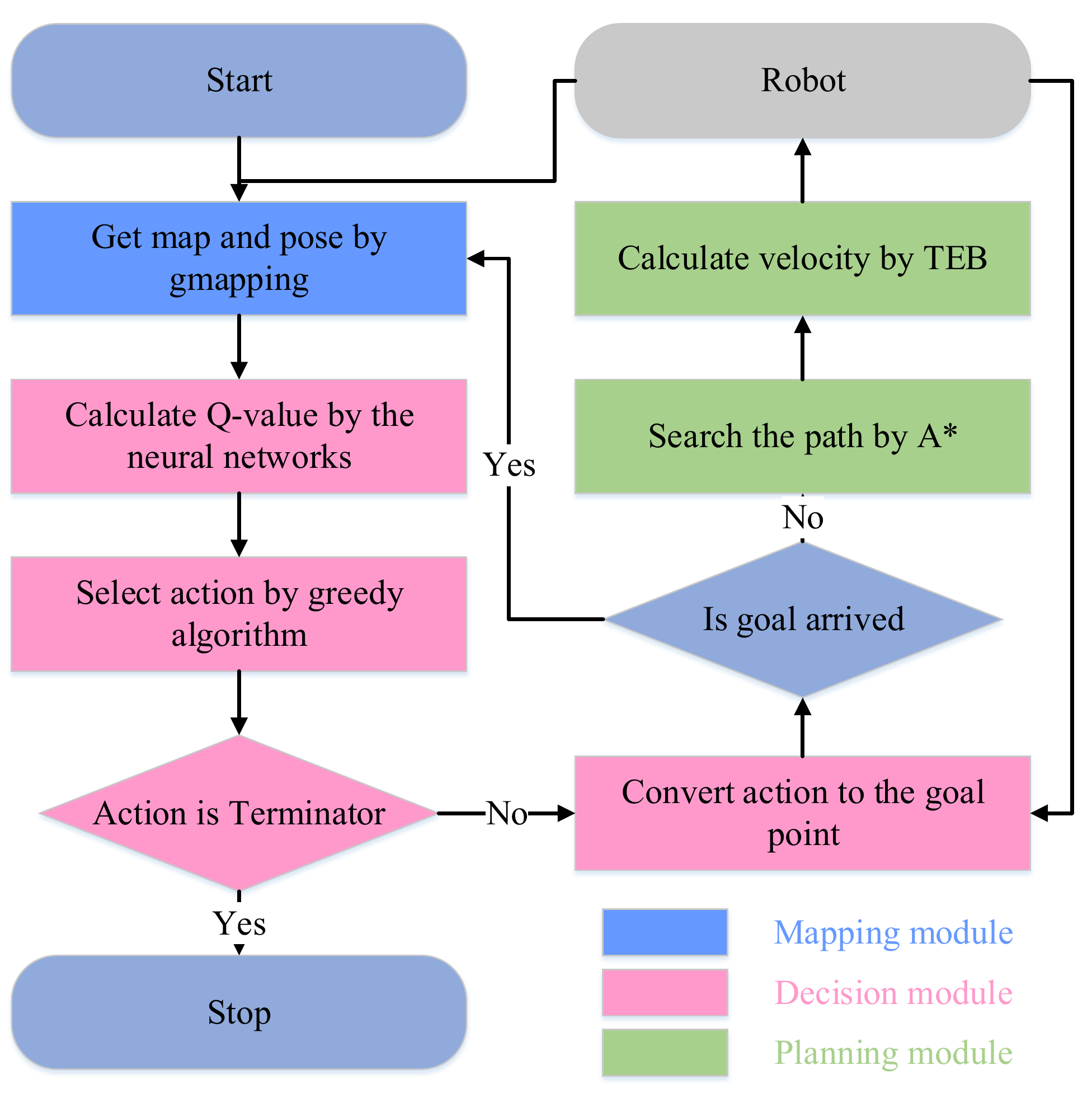}
  \caption{The diagram of automatic exploration system.}
  \label{Fig_flow}
\end{figure}

The work flow of the whole system is shown in Figure \ref{Fig_flow}. The robot collects odometry data and the point cloud data by scanning the environment and sends the data to the mapping module. In this paper, we apply the gmapping method as the mapping module which builds the map and obtains the pose in the built map. Based on the results of mapping module, we propose novel DRL-based decision algorithms(FCQN and AFCQN, where we use AFCQN as the example.) which utilize the built mapping and the localization as the input and gives the next goal point. As the global planner, $A^*$ algorithm receives the goal point from the decision module and the built map from the mapping module. It searches the feasible path which is composed of discrete points. The local planner translates the path to the velocity and sends to the robot. The robot moves around in the environment, receives the new sensor data and builds a new map. Then it begins the new process loop.

\section{Experiments}
In this section, we conduct experiments in the simulated world and the physical world to compare the performance of different algorithms. In our exploration framework, we present the DRL-based method as the decision algorithms. For the mapping module, we make use of graph-based methods, where the front-end is based on scan matching method\cite{olson2009real} and the back-end is based on graph optimization implemented by g2o\cite{kummerle2011g} package. The planning module is composed of $A^*$ search methods as the global planner and TEB as the local planner.
Firstly, we compare the different networks include original DQN, proposed FCQN and AFCQN with same state space and action space. Then we conduct experiments to find the similarity and difference between AFCQN and the frontier-based method. Finally, we deploy the algorithms on the real robot.

\subsection{Evaluation Metrics}
To compare the performance of different algorithms, we define the explored region rate, average path length and exploration efficiency.

Explored region rate embodies the completeness of the map built during exploration. It is defined as
\begin{equation}
\text{Explored region rate} = \frac{\text{Number of explored free cells}} {\text{Number of free cells in the real map}}.
\end{equation}

With the high explored region, we hope the agent move efficiently. Therefore, the average path length and exploration efficiency are introduced.
\begin{align}
\text{Average path length} &= \frac{\sum_i L{\tau _i}}{\text{Number of episodes}} \\
\text{Exploration efficiency} &= \frac{\sum_i (H(T)-H(0))_i}{\sum_i L{\tau_i}}
\end{align}
Where $L_{\tau _i}$ is the path length of the trajectory $\tau_i$ during $i$-th episode.

In general, a larger explored region rate leads to a longer path. So the larger explored region rate and shorter path are inconsistent. Exploration efficiency represents the entropy reduced per unit length. And it is a compromise metric for exploration.

\subsection{Simulation Setup}
\begin{figure}
  \centering
  \includegraphics[width=0.45\textwidth]{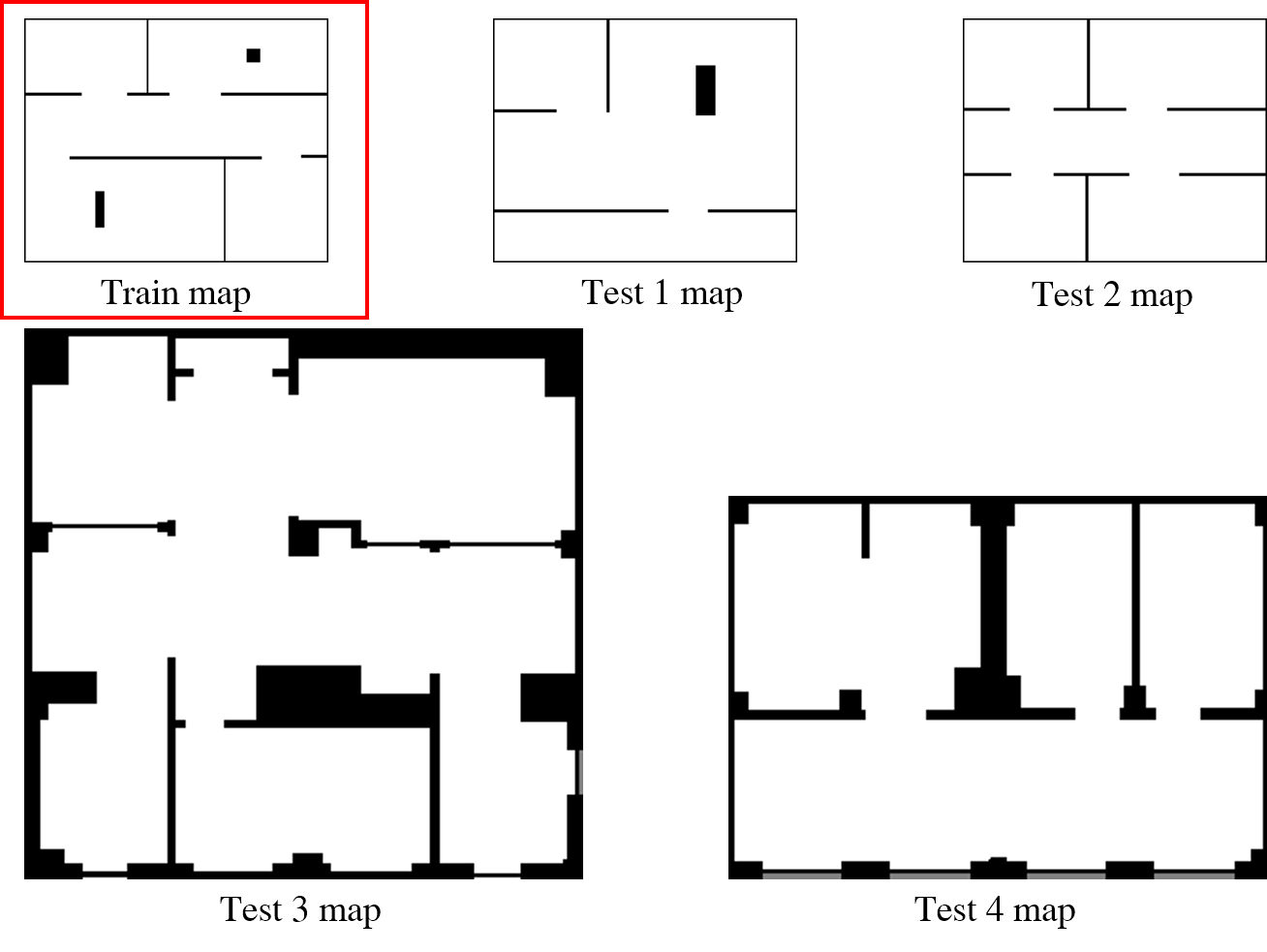}
  \caption{The maps for training and test.}
  \label{Fig_maps}
\end{figure}

We construct the simulation platform based on the Stage package\cite{gerkey2003player} of robot operation system(ROS) for agent training. We create a virtual map for algorithms training and design the maps with different layouts and sizes for testing. Figure \ref{Fig_maps} and the Table \ref{table_maps} show the details of the maps. The kinematic model of the agent is omnidirectional, which can go left or right without rotation.

\begin{table}
\centering
\caption{The details of the training and test maps.}
  \scalebox{1.2}{
   \begin{tabular}{l c c}
    \hline
    Map name  & Resolution & Real size(meter) \\
    \hline
    Training map  & 161$\times$201 & 5$\times$8 \\
    Test map 1 & 161$\times$201 & 5$\times$8 \\
    Test map 2 & 161$\times$201 & 5$\times$8 \\
    Test map 3 & 273$\times$277 & 13.65$\times$13.85 \\
    Test map 4 & 190$\times$267 & 9.5$\times$13.35 \\
    \hline
    \end{tabular}}
\label{table_maps}
\end{table}

\begin{figure}
  \centering
  \includegraphics[width=0.48\textwidth]{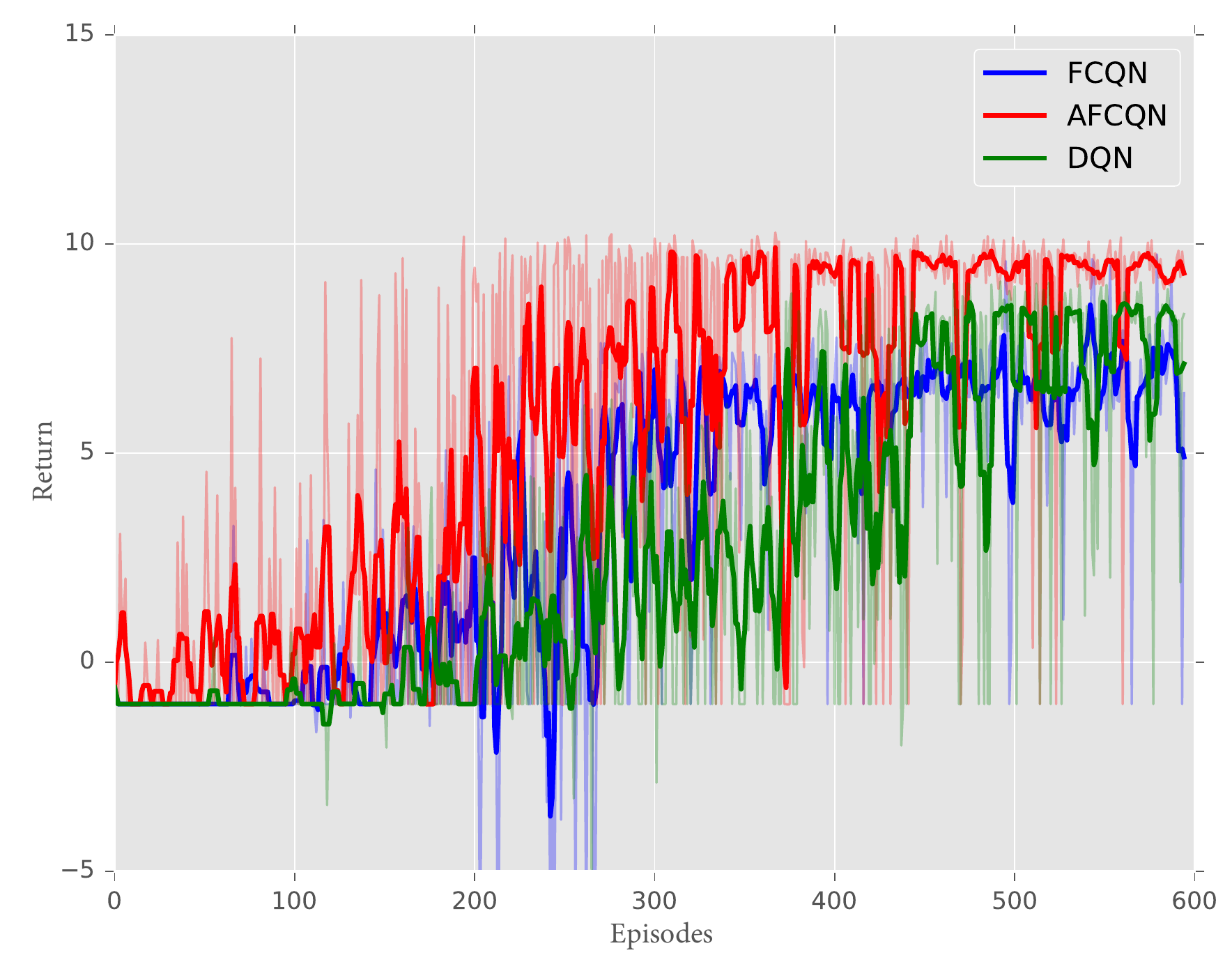}
  \caption{The return during training.}
  \label{Fig_train_score}
\end{figure}

\begin{figure}
  \centering
  \includegraphics[width=0.48\textwidth]{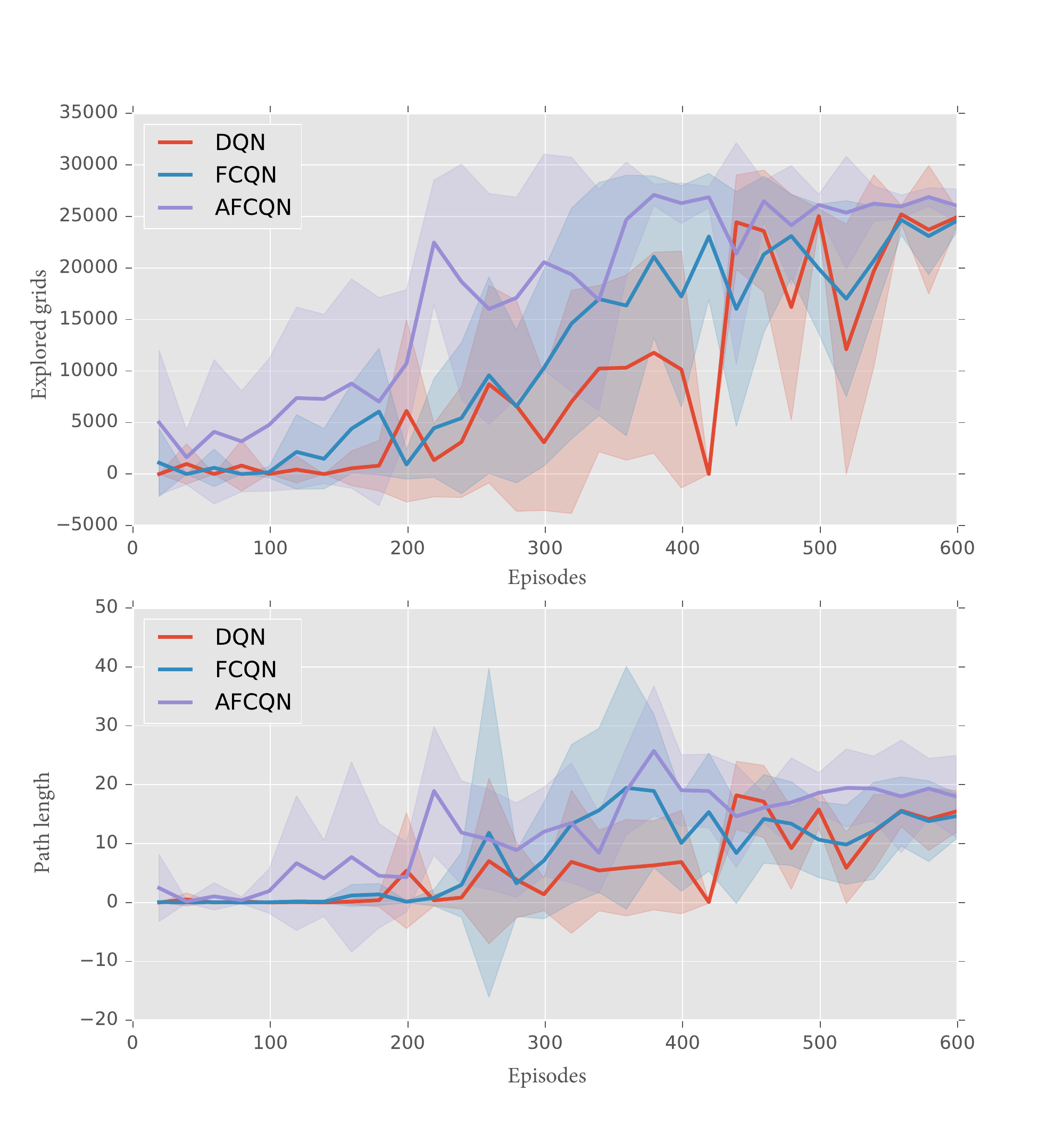}
  \caption{The explored grids and path length during training. The bold curves are average values over episodes. The shadow of each curve is corresponding to variance.}
  \label{Fig_free_path}
\end{figure}

\subsection{Training Analysis}
We conduct the training process on the training map. The inputs of the algorithms are the combined images including the grid map, the current position map, and the last position map. The position map draws the robot position on the blank map which has the same size as the built map.

Figure \ref{Fig_train_score} shows training processes. The returns grow during the training episodes and converging to the stable value. We can see that FCQN learns faster than DQN, but the final return is slightly smaller than DQN. The possible reason for this appearance is that DQN has more parameters than FCQN and better ability to remember complex decision sequences. Since AFCQN uses more information to find out the optimal parameters, AFCQN has a better performance in convergence speed and final return.

Figure \ref{Fig_free_path} describes the number of explored grids and the path length during the training episodes. From the pictures, we find out the edge segmentation task speeds up the exploration. AFCQN learns the safe strategy for movements and increases the path length quickly. However, the segmentation task also brings redundant movements which do not increase the new free cells. With the iteration of Q-values, the algorithm crops the redundant movements and achieves the better return.

\begin{table}
\centering
\caption{Results on the training map}
\scalebox{0.92}{
\begin{tabular}{l*5c}
\hline
Methods &   & Explored region rate & Path length & Explored efficiency \\
\hline
\multirow{2}{*}{AFCQN} & mean & \textbf{0.91} & 24.39 & 1226.12 \\
 & std & 0.0028 & 6.6152 & 311.0449 \\
\hline
\multirow{2}{*}{FCQN} & mean & 0.85 & \textbf{16.46} & \textbf{1634.00} \\
 & std & 0.0414 & 2.8107 & 297.4689 \\
\hline
\multirow{2}{*}{DQN} & mean & 0.84 & 23.19 & 1108.51 \\
 & std & 0.2190 & 5.9091 & 190.8171 \\
\hline
\end{tabular}}
\label{table_train}
\end{table}

\begin{figure}
  \centering
  \includegraphics[width=0.48\textwidth]{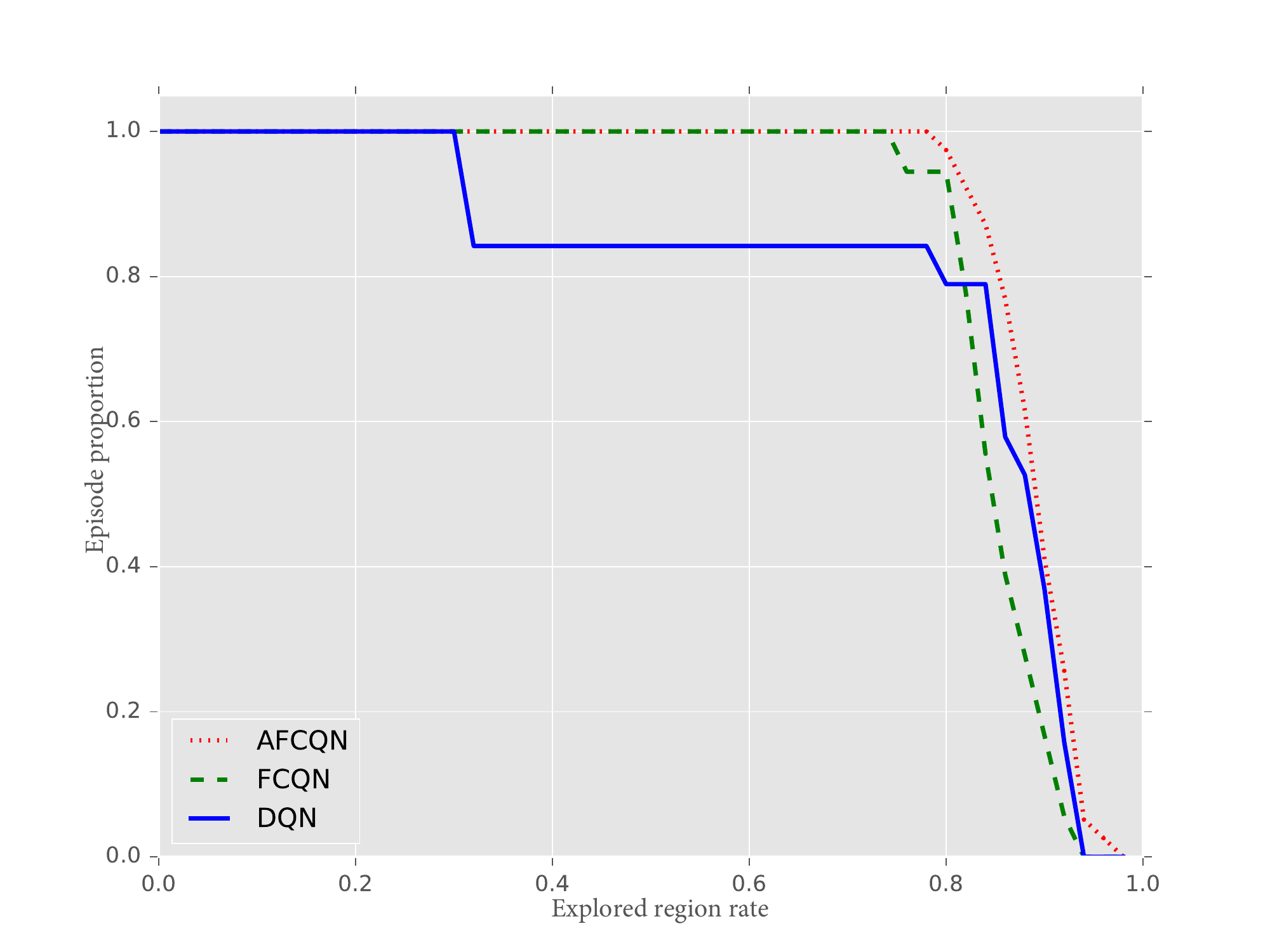}
  \caption{The episode proportion in trials goes with explored rate while exploration path length shorter than 25 meters.}
  \label{Fig_episode_rate}
\end{figure}

To compare the trained algorithms in the training map, we conduct 50 trials by selecting initial points randomly from free space.  Results are shown in Table \ref{table_train}. All algorithms finish the exploration completely on the training map. And AFCQN has highest explored region rate while FCQN has highest exploration efficiency.
The reason for high explored efficiency for FCQN is selecting the terminal action in time which avoids redundant movement. AFCQN has more attention on the completeness of the built map, which leads to the longer path. For most trials, DQN has a higher explored region rate than FCQN. But it fails in some cases. Figure \ref{Fig_episode_rate} shows the comprehensive performance. The curve means that the proportion of the episodes which achieve the specified explored region rate within 25 meters. According to the figure, we see AFCQN has a better performance than others. And FCQN has a stable performance, where its explored region rate distributes around 0.8. Except for failed cases, the explored region rate of DQN distributes around 0.9.

\subsection{Generalization Analysis}
\begin{figure}
  \centering
  \includegraphics[width=0.48\textwidth]{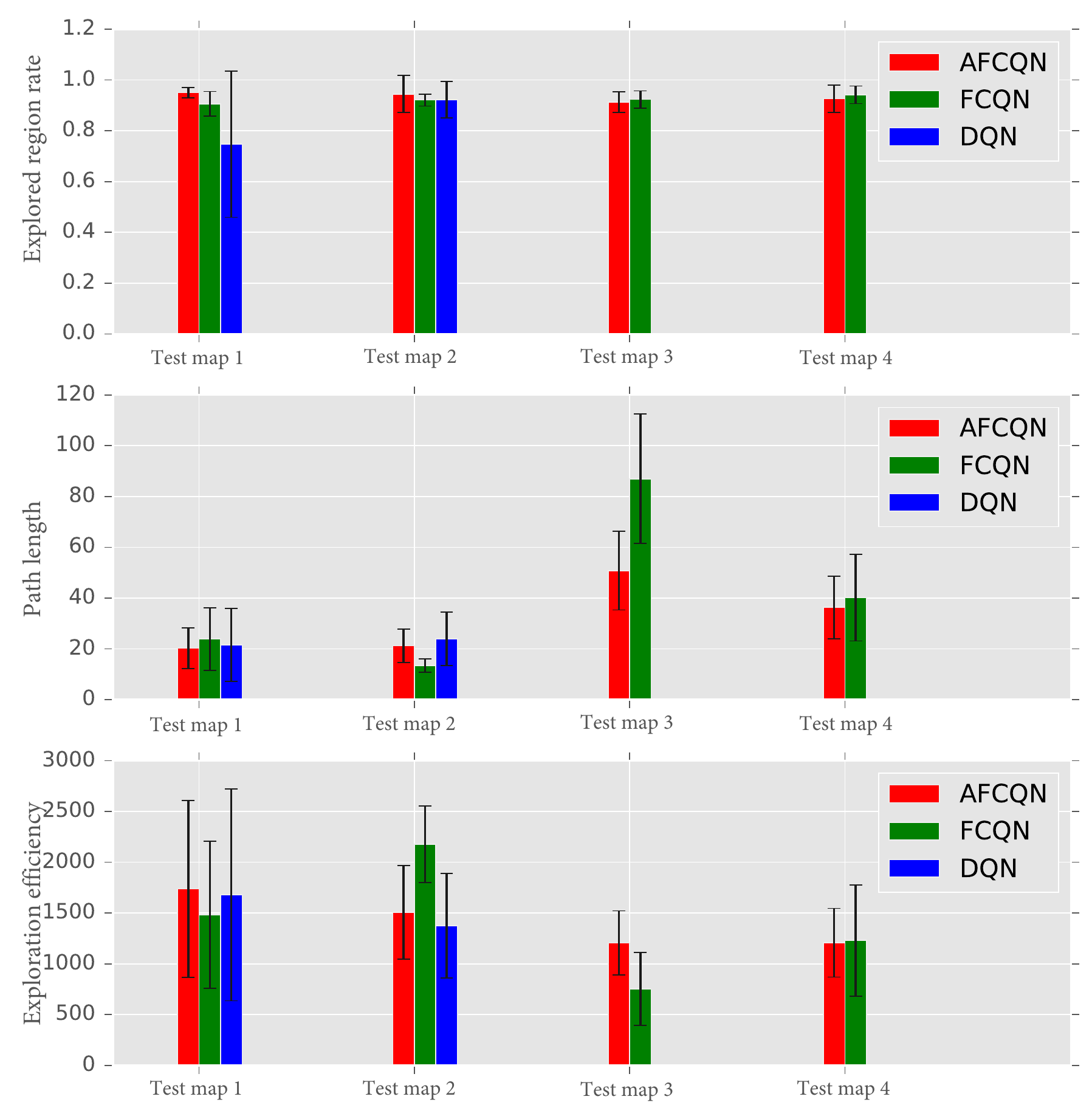}
  \caption{The statistic results of different algorithms in 4 test maps. Histogram means the average value for different metrics. And the error bar is the standard deviation.}
  \label{Fig_hist2d}
\end{figure}
In order to analyze the generalization of the algorithms, we design the test maps different from the train map. Considering the constraint of the stand DQN caused by the fully connected layers, we design two maps(test map 1 and test map 2) as shown in Figure \ref{Fig_maps} with the same size but different layouts with the train map. Our algorithms based on fully convolutional layer adapt well to variant map sizes. In addition, we design the two maps(test map 3 and test map 4) with different sizes with the train map. We test our algorithms on these maps. And the results are shown in Figure \ref{Fig_hist2d}.

According to Figure \ref{Fig_hist2d}, all algorithms have good generalization on the test maps which have different layouts from the train map. AFCQN has the highest explored region rate in test map 1 and FCQN follows. AFCQN and FCQN have the same explored region rate in test map 2 while FCQN has a less standard deviation. In addition, FCQN has a shorter path length. FCQN has a better performance than DQN. Compared to results in the training map, AFCQN has a higher explored region rate and the less explored region rate standard deviation than others in test maps.

Since the fully connected layers of DQN restrict its output dimension, DQN has poor adaptability for different size maps. Note that AFCQN and FCQN make use of fully convolutional layers instead of fully connected layers. They can address different size maps and adjust the output dimension adaptively with keeping a downsampling rate for generating action space. In this part, we compare the difference between AFCQN and FCQN. Although the layout and size are different from the train map, our proposed algorithms work well in new environments. AFCQN has a higher explored region rate. But FCQN has a higher exploration efficiency. The changes of layout and scale increase the standard deviation of explored region rate. 

From the above two groups of experiments, AFCQN and FCQN are better than DQN in all maps.
The layout of the map has an influence on the algorithms.
Comparing the test map 1 and test map 2, we can find the common feature between the training map and test map 2. That is they both have many rooms and ``doors''. And the layouts of the test map 1 and 3 are relatively spacious. With this special feature, FCQN has better performance in exploration efficiency. The reason for this result is that FCQN can end the exploration in time and has shorter exploration path. Note that we add the edge segmentation as the auxiliary task, AFCQN focus on the completeness of the built map(since AFCQN has higher explored region rate) and has longer exploration path which leads to lower exploration efficiency.
In spacious environments such as test map 1 and test map 3, the exploration strategy is simple than in crowded environments such as test map 2 and training map. This means that it is easier to complete exploration. And AFCQN can detect the completeness of the built map and end the process in time. FCQN need more step to confirm the finish. These results show that AFCQN has better ability to detect the completeness of the built map.


\subsection{Relationship to Frontier-based Methods }
\begin{figure*}
\centering
  \includegraphics[width=0.9\textwidth]{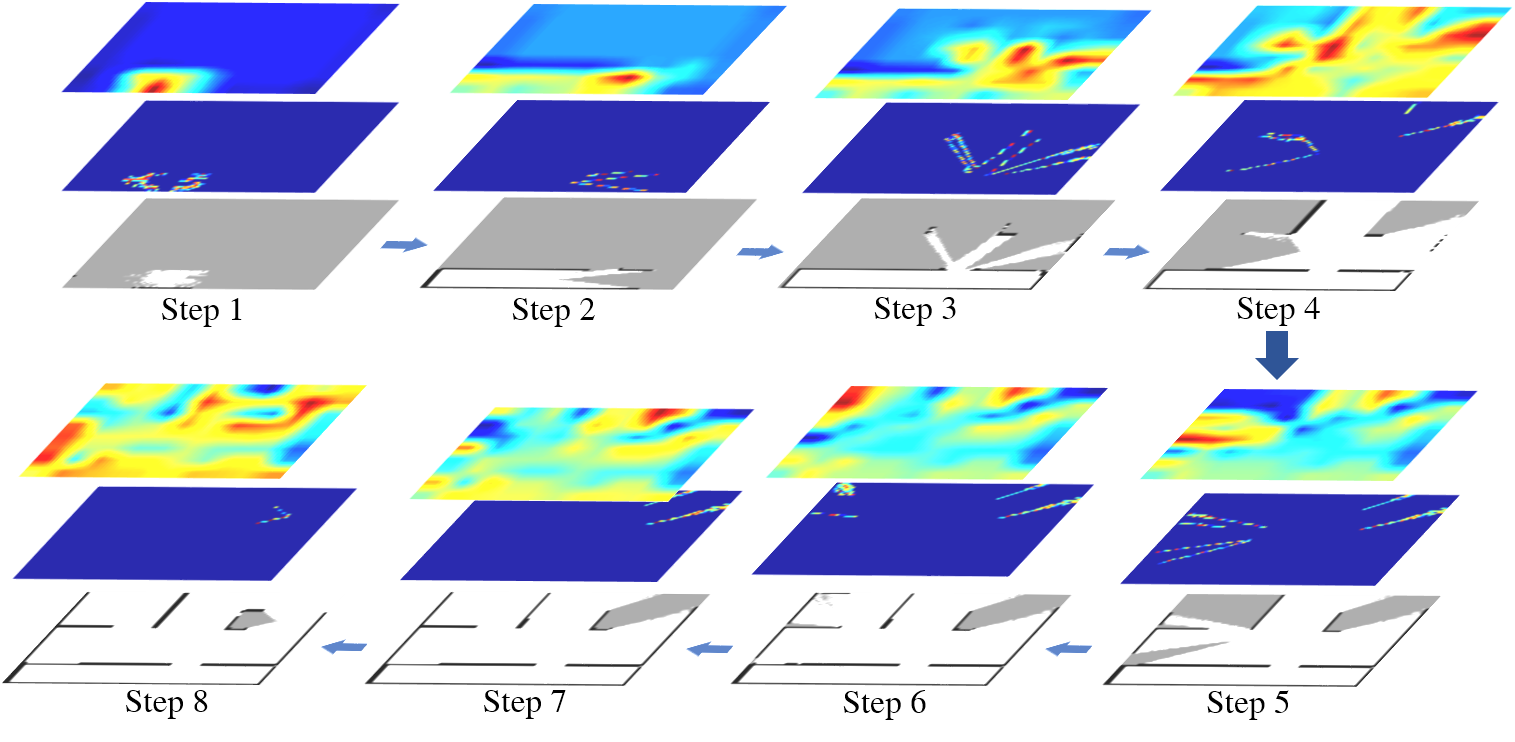}
  \caption{The decision process of AFCQN during exploration. The whole process composed of 8 steps. Each step has 3 layers including built occupancy grid map(down), frontier(middle), Q-value map(top). After the 8th step, AFCQN selects the terminal action and stops exploration.}
  \label{Fig_frontier}
\end{figure*}

\begin{figure}
  \centering
  \includegraphics[width=0.48\textwidth]{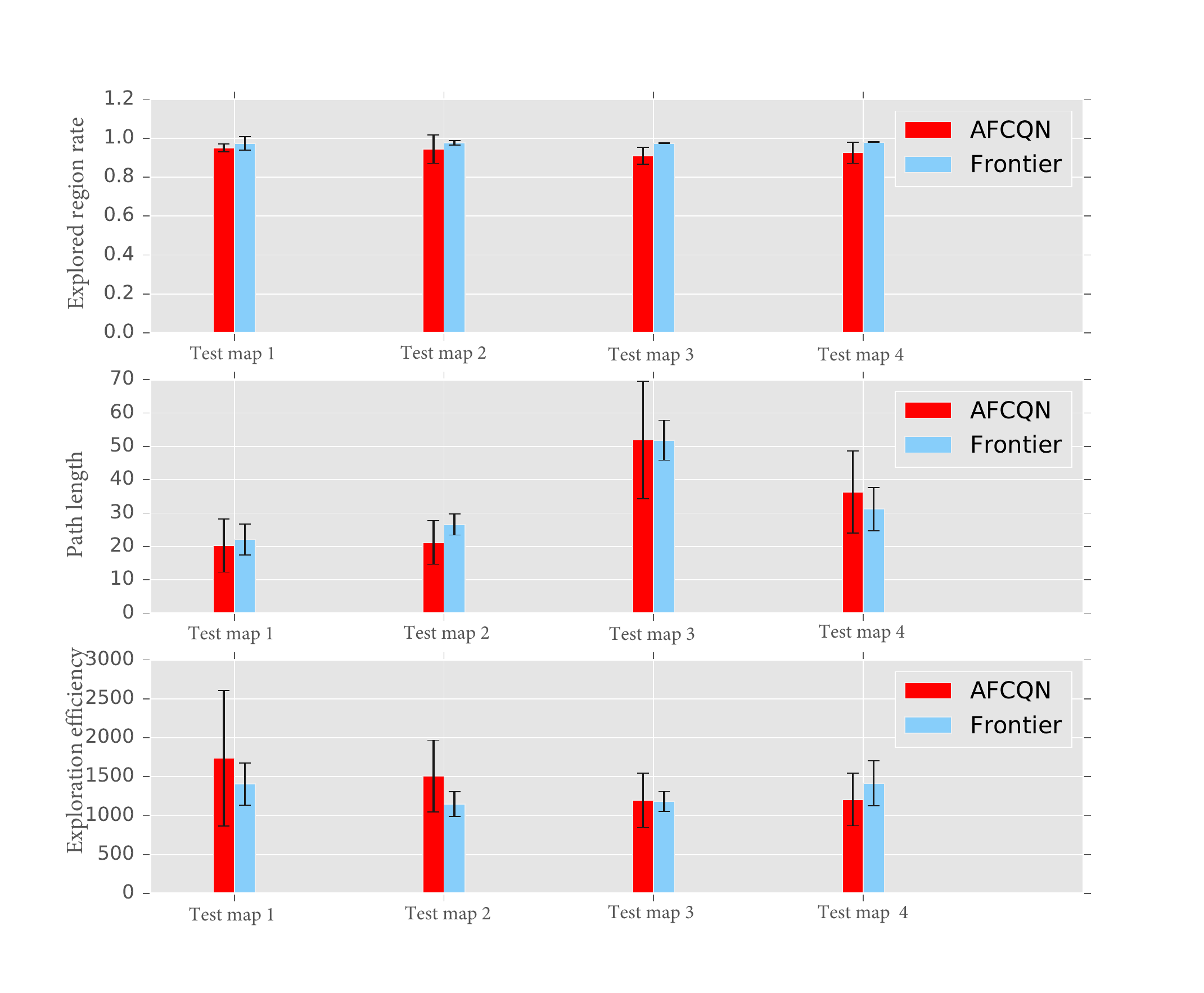}
  \caption{The statistic results of AFCQN and frontier-based method in 4 test maps. Histogram means the average value for different metrics. And the error bar is the standard deviation.}
  \label{Fig_hist2d_frontier}
\end{figure}

\begin{figure}
  \centering
  \includegraphics[width=0.48\textwidth]{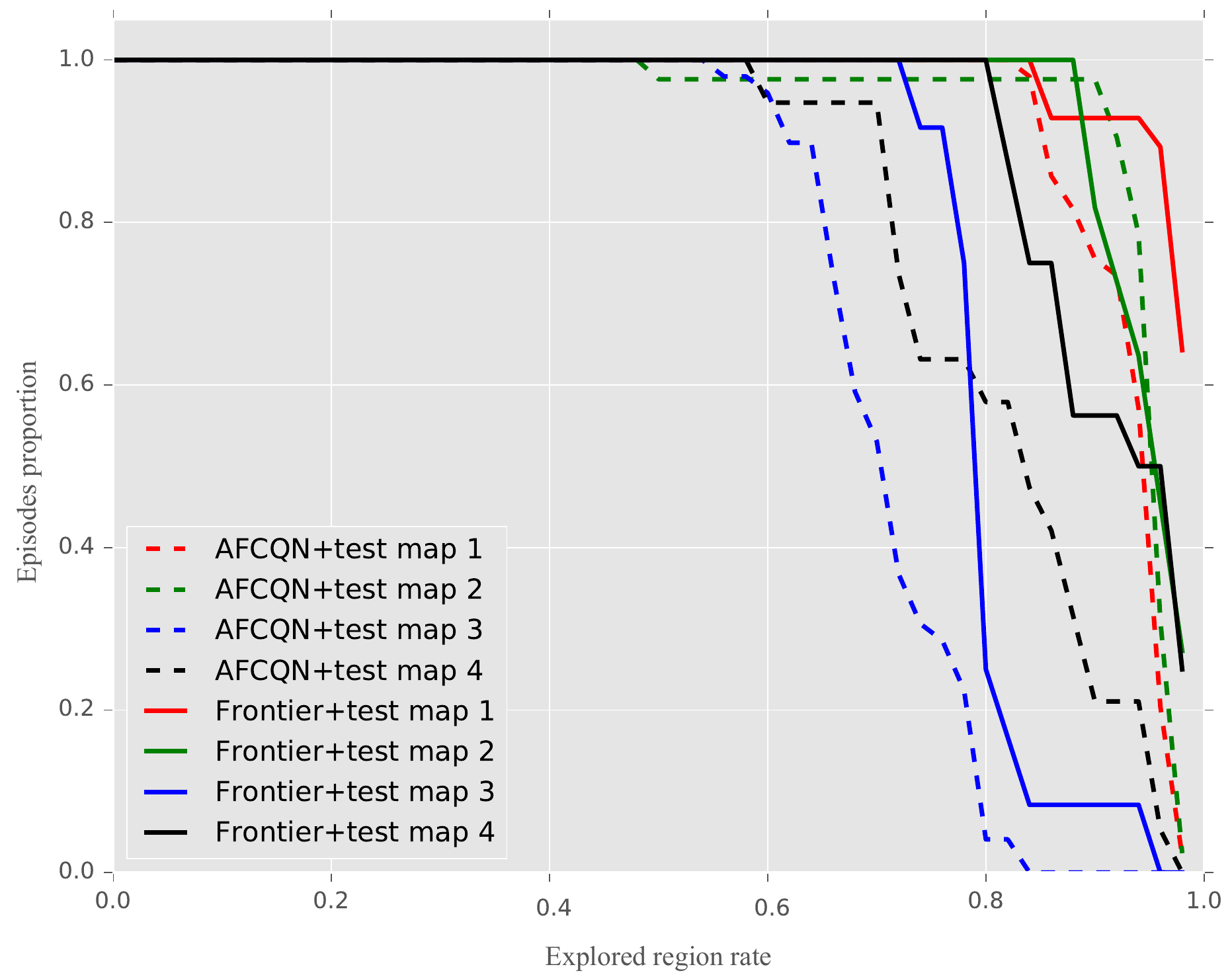}
  \caption{The episodes proportion in trials goes with explored rate.}
  \label{Fig_afcqn_frontier}
\end{figure}

\begin{figure}
  \centering
  \includegraphics[width=0.35\textwidth]{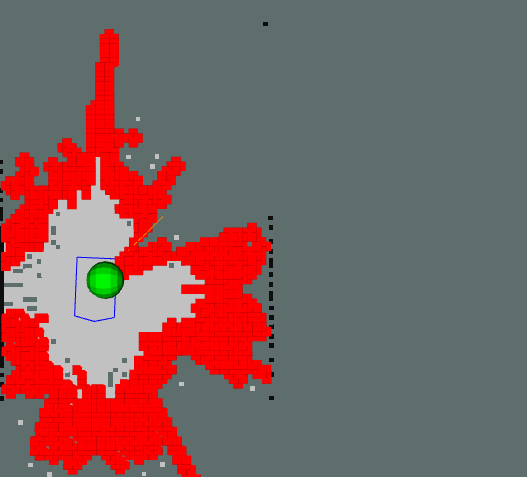}
  \caption{The frontier-based methods fail case. The blue polygon is the base of the robot. The red grids are frontiers. The green ball is the center of the frontiers. In this case, since the robot position and frontiers center overlap, the exploration stops. }
  \label{Fig_frontier_failed}
\end{figure}

Since AFCQN takes the edge segmentation as the auxiliary task, we further analyze how this task affects the decision of AFCQN.
In the frontier-based methods, the frontiers which are the boundaries between free space and unknown space are the decision candidates,
so we compare the decision process of AFCQN to frontier based methods. Figure \ref{Fig_frontier} describes the process.

In the early decision stage, the valuable decision positions distribute around the frontiers. With expanding of the map, the frontiers speed out. AFCQN learns intuition for maximizing the increasing free cells. In step 5, the algorithm selects the center of frontiers in the left part of the map. It is more efficient than selecting a center of any frontier. In step 6, there are two frontiers on the map. The upright frontier is the candidate which increases more free grids. The upleft frontier will increase a few free grids but closer to the current position. It can be seen that AFCQN selects the upleft point firstly and the upright point secondly. This strategy is better than selecting the upright firstly for considering the completeness and the whole path length. After step 8, AFCQN selects the terminal action to stop exploration.

We also compare AFCQN to the frontier-based method. Figure \ref{Fig_hist2d_frontier} shows the results on the 4 test maps. The frontier-based method achieves a higher explored region rate. Since the decision point of this method is selected from frontier centers, it explores the environment continually until there are no reachable points. In other words, it usually has a long path. Although AFCQN is inspired by the built map edges, it estimates the status of the whole maps in order to select the terminal action in time. Therefore, AFCQN has higher exploration efficiency in the first three test maps.

Figure \ref{Fig_afcqn_frontier} presents the stability of the algorithms. With constraints of the distance, the frontier-based method is better than AFCQN. However, the frontier-based method also faces with some special cases which make exploration fail. We explain one case in Figure \ref{Fig_frontier_failed}. Lidar receives the point cloud and renders to grid map. The frontier-based method detects the frontiers shown as red grids and calculates the center represented by a green ball. Unfortunately, the robot is standing at this center point, then it stops the exploration. Since the inputs of AFCQN include the robot current position and last decision position, this case never happens during conducting AFCQN method.

\subsection{Physical World Experiments}
\begin{figure}
  \centering
  \includegraphics[width=0.48\textwidth]{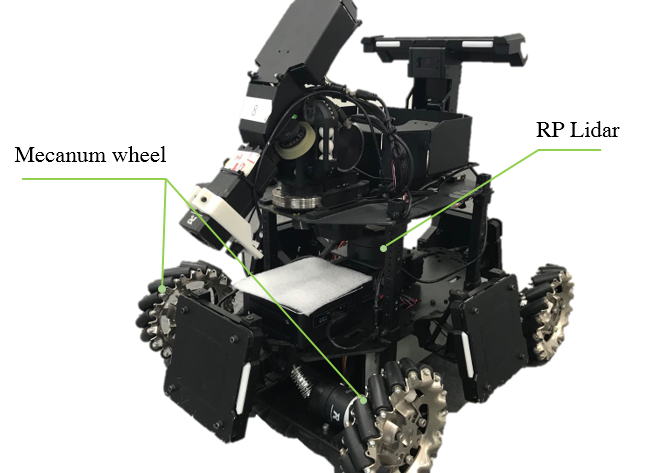}
  \caption{The hardware of the DJI Robomaster platform.}
  \label{Fig_hardware}
\end{figure}

Since our algorithm output is the decision point instead of the robot control, it is easier to transfer to the physical robot. We employ the DJI Robomaster robot platform shown in Figure \ref{Fig_hardware}. The major components include the gimbal and the chassis. The chassis equips with mecanum wheel for omnidirectional movements. We fixed an RPLidar which has similar properties with simulated LiDAR on the chassis for mapping. Then, we test our algorithm with the platform in the real environment.

\begin{figure}
  \centering
  \includegraphics[width=0.48\textwidth]{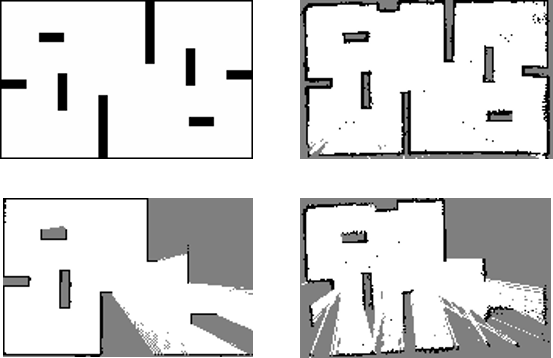}
  \caption{The simulation map(upleft) and the real built map(upright). The downleft image is the observation (built map) coming from the simulation environment and the downright image comes from the real-world environment.}
  \label{Fig_built_map}
\end{figure}
\begin{figure}
  \centering
  \includegraphics[width=0.48\textwidth]{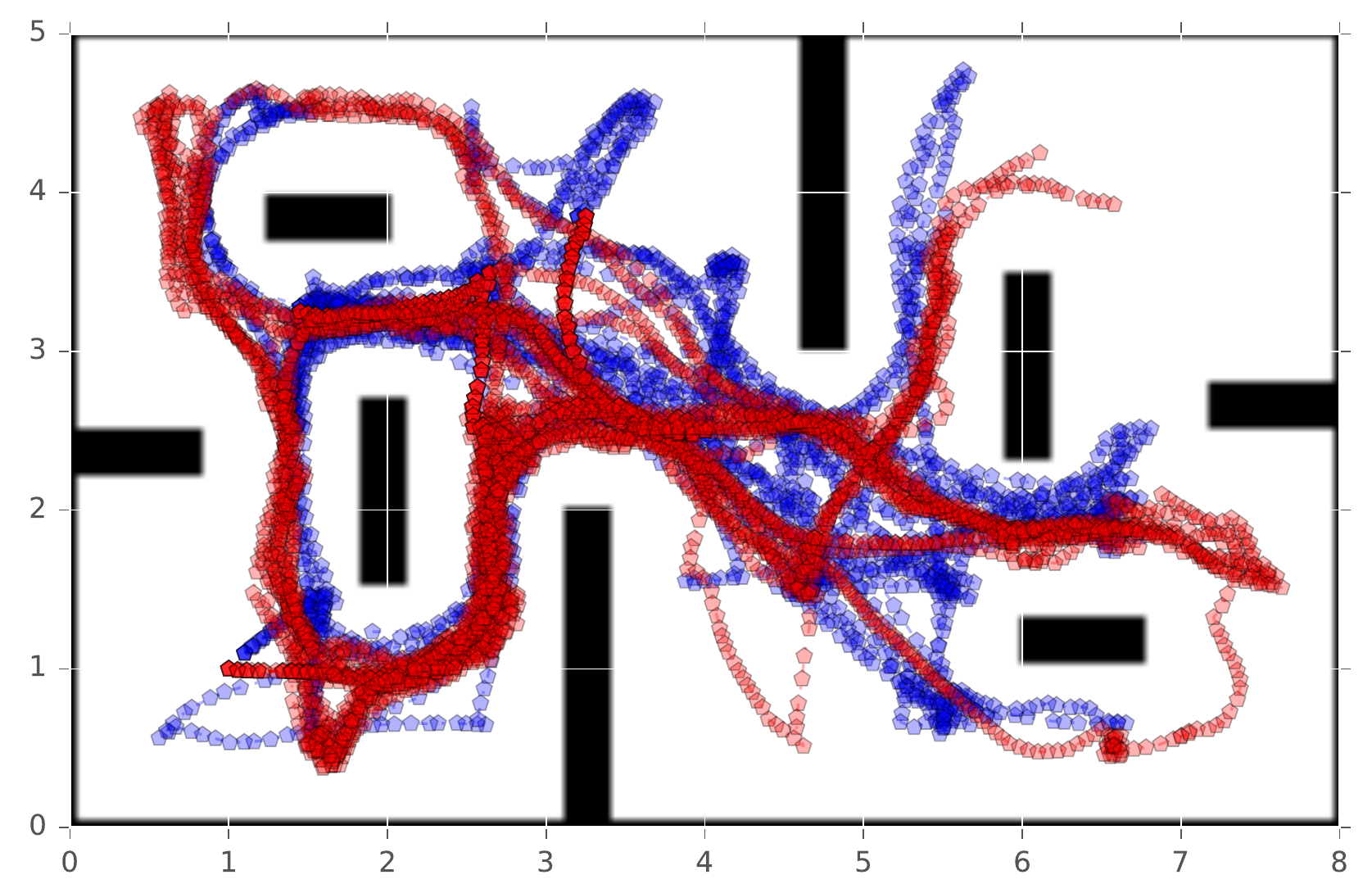}
  \caption{The exploration trajectories of AFCQN in trials. The red trajectories come from simulation experiments while the blue trajectories come from real experiments.}
  \label{Fig_real_sim_traj}
\end{figure}

Here, we show the results of the physical world experiment to the simulated world. The simulated map and built real world map are shown in Figure \ref{Fig_built_map}. Due to the difference in sensor errors and environments, the observations of their experiments are different. Table \ref{table_icra} shows the comparing results.
Frontier-based methods have highest explored region rate in both the physical world and simulated world. However, our proposed method AFCQN has best explored efficiency, since our method monitor the completeness of the built map and stop the exploration in time.
It is obvious that the distinction between the two observations impacts the explored region rate.
DQN is sensitive of inputs and  fails to explore the physical world for several times. For other methods,
the variances of the explored region rate are same. Those differences have no effect on exploration path length. The variance of the path length is decided by the local planner. And we have different local planner parameters for simulation and real experiments in order to adapt the different kinematic models.
To explain AFCQN behaviour, the exploration trajectories of AFCQN in the trails are shown in Figure \ref{Fig_real_sim_traj}. The most trajectories of the real experiments overlap with those of the simulation experiments. And the results imply that the decision processes of the two experiments are similar.

\begin{table}
\centering
\caption{Results of real world and simulation experiments}
\scalebox{0.75}{
\begin{tabular}{l*6c}
\hline
Environment & Methods  &  & Explored region rate & Path length & Explored efficiency \\
\hline
\multirow{8}{*}{Real} & \multirow{2}{*}{AFCQN} & mean & 0.84  & \textbf{14.76}  & \textbf{863.82} \\
 					  &						   & std & 0.0344 & 3.9634 & 227.0628 \\
 					  \cline{3-6}
					  &\multirow{2}{*}{FCQN} & mean & 0.78   & 16.18  & 796.37 \\
 					  &						 &  std&  0.0543 & 4.2317 & 214.7032 \\
 					  \cline{3-6}
 					  & \multirow{2}{*}{DQN} &  mean &  0.54   &  6.87  & 410.21 \\
 					  &						  				   & std & 0.4173 &5.8931 & 340.1092 \\
 					  \cline{3-6}
 					  & \multirow{2}{*}{Frontier}& mean & \textbf{0.91}   & 19.23  & 840.79 \\
 					  &						   			 			& std & 0.0134 & 2.0172 & 167.3728 \\
\hline
\multirow{8}{*}{Sim}  & \multirow{2}{*}{AFCQN}	& mean & 0.92 & \textbf{15.78} & \textbf{986.95} \\
 					  &							& std & 0.0325 & 6.2216 & 347.7612 \\
 					  \cline{3-6}
 					  & \multirow{2}{*}{FCQN}	& mean & 0.87 & 16.58 & 925.74 \\
 					  &											& std & 0.0473 & 8.3158 & 314.8210 \\
 					  \cline{3-6}
 					  & \multirow{2}{*}{DQN}	&mean & 0.81 & 17.23 & 806.94 \\
 					  &											& std & 0.1917 & 9.0315 & 425.1479 \\
 					  \cline{3-6}
 					  & \multirow{2}{*}{Frontier}	& mean & \textbf{0.95} & 17.19 & 962.34 \\
 					  &							   					& std & 0.0127 & 2.1475 & 185.6325 \\
 					
\hline
\end{tabular}}
\label{table_icra}
\end{table}

%

\section{Conclusion}
In this paper, we construct a novel framework for robot exploration based on mapping, decision, and planning modules. And we declare the general model for each module. Under this framework, each module is mutual independent and can be implemented via various methods.

The key contribution of this paper is proposing a decision algorithm based on DRL which takes the partial map as the input. In order to improve the training efficiency and generalization performance, we attach the edge segmentation task to train the feature-extracting networks auxiliary.

Since the proposed framework is combined with the existing navigation algorithms, the algorithms based on the framework learn faster than the end-to-end methods\cite{tai2016mobile}\cite{mirowski2016learning} which undergo thousands of episodes.
The experiments show that AFCQN has better generalization performance in different maps compare to end-to-end methods and stand DQN.
Compared to the frontier-based method, AFCQN can avoid failed case and make use of properties of the sensor to explore the environment more efficiently, which makes the decision module look more intelligent.
The experiments in the physical world show that our algorithm is easy to transfer from the simulator to the physical world with a tolerable performance.

However, there are some further studies. Since the gridding map is used to generate discrete actions, the final decision sequence based on the discrete action space is suboptimal. Hilbert occupancy map\cite{ramos2016hilbert} is a potential substitute, which is used to find out the optimal exploration path in a given map. How to combine this map and policy gradient method is a valuable topic.
Deep recurrent neural networks can be adapted to improve the online learning performance\cite{ergen2018efficient}, and it can be introduced to this task to augment the memory of the algorithm. In addition, how to embed the available auxiliary information and domain knowledge into the DRL method is also a very meaningful research topic for automatic exploration.

\ifCLASSOPTIONcaptionsoff
  \newpage
\fi



%
\bibliographystyle{IEEEtran}
\bibliography{IEEEabrv,auto_exploration}

%

%
%
%




\end{document}